\documentclass[conference]{IEEEtran}
\IEEEoverridecommandlockouts
\usepackage{cite}
\usepackage{amsmath,amssymb,amsfonts}
\usepackage{algorithmic}
\usepackage{graphicx}
\usepackage{textcomp}
\usepackage{xcolor}
\usepackage[a4paper, total={184mm,239mm}]{geometry}

\usepackage{subfigure}
\usepackage[algo2e,ruled,vlined,linesnumbered]{algorithm2e}
\usepackage{booktabs}       % professional-quality tables
\usepackage{multirow}

\def\BibTeX{{\rm B\kern-.05em{\sc i\kern-.025em b}\kern-.08em
    T\kern-.1667em\lower.7ex\hbox{E}\kern-.125emX}}
\begin{document}

% \title{Calibrated BatchNorm: Improving Robustness Against Noisy Weights in Neural Networks\\
% % {\footnotesize \textsuperscript{*}Note: Sub-titles are not captured in Xplore and
% % should not be used}
% % \thanks{Identify applicable funding agency here. If none, delete this.}
% }

\title{Robust Processing-In-Memory Neural Networks via Noise-Aware Normalization\\
}

% \author{
% \IEEEauthorblockN{Anonymous submission}
% }

\author{\IEEEauthorblockN{Li-Huang Tsai}
\IEEEauthorblockA{\textit{Department of Computer Science} \\
\textit{National Tsing-Hua University}\\
Hsinchu, Taiwan \\
lihuangtsai@gapp.nthu.edu.tw}
\and
\IEEEauthorblockN{Shih-Chieh Chang}
\IEEEauthorblockA{\textit{Department of Computer Science} \\
\textit{National Tsing-Hua University}\\
Hsinchu, Taiwan \\
scchang@cs.nthu.edu.tw}
\and
\IEEEauthorblockN{Yu-Ting Chen}
\IEEEauthorblockA{\textit{Google Research} \\
\textit{Google}\\
CA, USA \\
yutingchen@google.com}
\and
\IEEEauthorblockN{Jia-Yu Pan}
\IEEEauthorblockA{\textit{Google Research} \\
\textit{Google}\\
CA, USA \\
jypan@google.com}
\and
\IEEEauthorblockN{Wei Wei}
\IEEEauthorblockA{\textit{Google Research} \\
\textit{Google}\\
CA, USA \\
wewei@google.com}
\and
\IEEEauthorblockN{Da-Cheng Juan}
\IEEEauthorblockA{\textit{Google Research} \\
\textit{Google}\\
CA, USA \\
dacheng@google.com}
}

\maketitle

\begin{abstract}
% This document is a model and instructions for \LaTeX.
% This and the IEEEtran.cls file define the components of your paper [title, text, heads, etc.]. *CRITICAL: Do Not Use Symbols, Special Characters, Footnotes, 
% or Math in Paper Title or Abstract.
Analog computing hardwares, such as Processing-in-memory (PIM) accelerators, have gradually received more attention for accelerating the neural network computations. However, PIM accelerators often suffer from intrinsic noise in the physical components, making it challenging for neural network models to achieve the same performance as on the digital hardware. Previous works in mitigating intrinsic noise assumed the knowledge of the noise model, and retraining the neural networks accordingly was required.   
In this paper, we propose a noise-agnostic method to achieve robust neural network performance against any noise setting. Our key observation is that the degradation of performance is due to the \textit{distribution shifts} in network activations, which are caused by the noise.  To properly track the shifts and calibrate the biased distributions, we propose a ``noise-aware'' batch normalization layer, which is able to align the distributions of the activations under variational noise inherent in the analog environments.  Our method is simple, easy to implement, general to various noise settings, and does not need to retrain the models.
We conduct experiments on several tasks in computer vision, including classification, object detection and semantic segmentation. The results demonstrate the effectiveness of our method, achieving robust performance under a wide range of noise settings, more reliable than existing methods.  We believe that our simple yet general method can facilitate the adoption of analog computing devices for neural networks.

% Analog computing hardware has gradually received more attention by the researchers for accelerating the neural network computations in recent years. However, the analog accelerators often suffer from the undesirable intrinsic noise caused by the physical components, making the neural networks challenging to achieve ordinary performance as on the digital ones. We suppose the performance drop of the noisy neural networks is due to the distribution shifts in the network activations. In this paper, we propose to recalculate the statistics of the batch normalization layers to calibrate the biased distributions during the inference phase. Without the need of knowing the attributes of the noise beforehand and any further re-training procedure, our approach is able to align the distributions of the activations under variational noise inherent in the analog environments. In order to validate our assumptions, we conduct quantitative experiments and apply our methods on several computer vision tasks, including classification, object detection, and semantic segmentation. The results demonstrate the effectiveness of achieving noise-agnostic robust networks and progress the developments of the analog computing devices in the field of neural networks.

\end{abstract}

\begin{IEEEkeywords}
Deep neural networks, processing-in-memory, noise mitigation
\end{IEEEkeywords}

\section{Introduction}
\label{sec::intro}

The recent success of deep neural networks has raised the interest in discovering suitable hardware devices for neural network inference which demands computational resources and energy consumption heavily. As the deployment of network models becoming widely spread on a variety of edge devices, it is urgent to design hardware to satisfy the needs of power consumption and performance. In addition to the widely-used digital circuits (e.g. GPU) which have already been well developed, analog computing has attracted more attention in recent years since non-volatile memory devices are favourable in accelerating the inference of neural networks~\cite{ISAAC, Analog_or_Digital_Approach}. In comparison to the digital platforms, processing-in-memory (PIM) analog computing has demonstrated orders of speed acceleration and lower power consumption, allowing it to become a reasonable choice for the neural network inference.

\begin{figure*}[!t]
\centering
\subfigure[KL Divergence]
{
\label{fig:kl_div}
\includegraphics[width=0.47\textwidth]{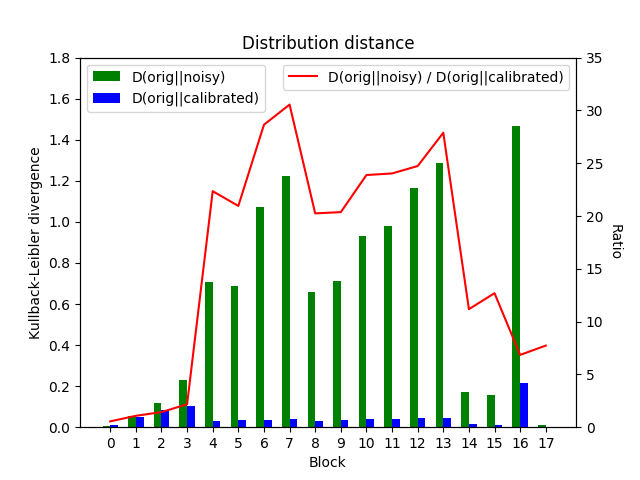}
}
\subfigure[JS Divergence]
{
\label{fig:js_div}
\includegraphics[width=0.47\textwidth]{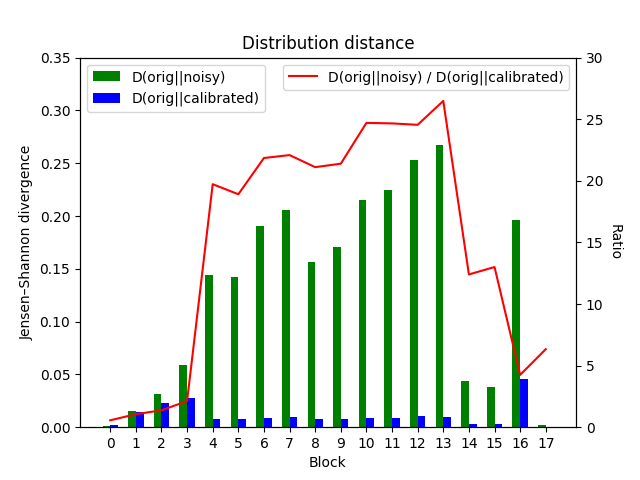}
}

\caption{Distribution distance between the activations in ResNet-34 before and after the MUL type noise injection. The blue and the green bars represent the divergence of the activations with and without applying our approach, respectively, while the red line illustrates the ratio between them. It can be observed that the distributions of the intermediate presentation with noise injected are far from the ones without any noise while the BatchNorm statistics are not calibrated, which results to the degradation in the final performance. In contrast, our approach calibrates the BatchNorm statistics and shifts the distributions close to the clean ones in both (a) KL divergence and (b) JS divergence.}
\label{fig:distribution_distance}

\end{figure*}

However, in analog computation, intrinsic noise in the physical components of the device can affect the computation and result in intolerable performance drop of the neural networks, making it impractical as a replacement of the digital circuits.
%despite the advantages of analog computing in speed and power consumption.
%The performance degradation is due to the undesirable intrinsic noise inherent in the physical components of the analog devices. 

There have been several works proposed to address the issue of intrinsic noise in analog computation. \cite{qin2018training}, for example, proposed to fine-tune the neural networks after the conventional training phase. Model weights were injected with Gaussian noise to simulate the analog computing scheme while fine-tuning. The proposed noise injection training allowed the networks to be robust to the noise on the analog devices during the inference. However, this kind of noise injection training requires a prior knowledge of the attributes of the noise on the target device, which it is impractical to acquire beforehand.
% Moreover, simulating the injected noise based on the prior knowledge could be unrealistic for the target device.
Each of these noisy-trained networks is specifically fit to a certain noise scale and further fine-tuning is required for different ones. The process of fine-tuning is inefficient as it demands additional training time and computing resources on top of the original training phase. As a result, the inefficiency and the unaffordable cost of noise injection training make it unsatisfactory for facilitating the analog computing in practice.

The natural question that arises is how does the noise on analog devices cause the performance drop in the inference of neural networks? Our key idea is that the noise shifts the distribution of the model activations away from the original one without any noise, and such distribution shift therefore causes the performance drop in the neural networks. To verify our hypothesis, we measure and compute the distance between the distribution of clean activations and that of noisy activations. In Figure~\ref{fig:distribution_distance}, the green bars show the distance between the two distributions. In this analysis, we consider two distance functions, namely, Kullback-Liebler and Jensen-Shannon divergence. It is obvious that the noisy activations are significantly disturbed by the noise and are shifted far away from the clean ones. 

To mitigate such distribution shift on the activations, we propose a method to rectify the disturbed activation distributions, more importantly, without the need of any prior knowledge of the noise. 
Our method exploits the characteristics of the BatchNorm layer~\cite{ioffe2015batch} which is capable of normalizing the mismatched distributions of the activations among mini-batches.  We propose a novel adaption to the BatchNorm layer, extending its ability in alleviating the distribution shift in the presence of noise.
Specifically, we observe that the noise would make the running estimates of mean and variance (maintained in the BatchNorm layers) inaccurate, which significantly weakens the normalization effect of BatchNorm layers. We propose a noise-aware calibration on the calculations of those running statistics, and we show that these noise-aware calibrated statistics are able to effectively normalize the mismatched distributions, as depicted by the blue bars in  Figure~\ref{fig:distribution_distance}. 

Our proposed method has several advantages. Unlike the noise injection training which requires additional resource for fine-tuning, the cost of our method that merely keeps track of the running estimates is negligible. Furthermore, our method does not require prior knowledge of the noise, and can adapt to various scales of noise. In fact, our method continuously tracks the effect of noise, employing the proposed Algorithm~\ref{alg:act} to calculate the running mean and variance of the model activations. Therefore, our method is a practical approach to mitigate the noise interference in the neural network inference on analog computing.

We validate the performance of our method with a variety of computer vision tasks, including image classification, object detection and semantic segmentation. Our approach is able to alleviate the disturbing noise and improve the network robustness against a wide range of noise scales. 
The contributions of this paper are summarized as follows:

% Additionally, we analyze the effectiveness of our approach employing on several representative models under variational noise in a number of experiments quantitatively and qualitatively. 

\begin{itemize}
    \item We propose a noise-aware calibration in BatchNorm statistics, which effectively rectifies the shifted distribution caused by the noise during analog computing.
    \item Our approach requires negligible additional cost for calibrating the BatchNorm statistics, comparing to the unaffordable cost required in the noise injection training.
    \item Our approach is adaptive to variational noise and needs merely a few adjustments for different scales of noise.
    \item The effectiveness, efficiency and simplicity of our noise-aware method can facilitate the development of the analog computing and its deployment into practice.
\end{itemize}

\section{Related Work}
 \label{sec::related_work}

\begin{figure*}[!t]
\centering
\includegraphics[width=0.90\textwidth]{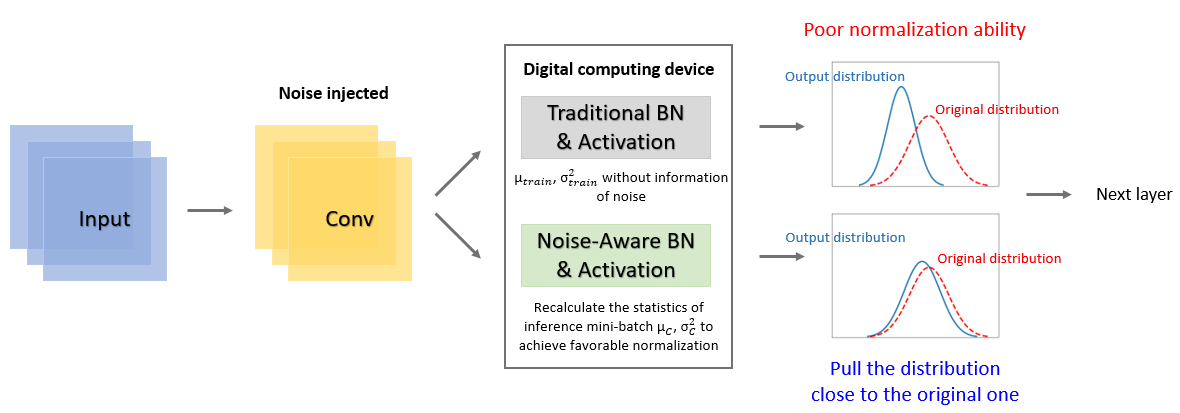}
\caption{The schematic diagram of ``Noise-Aware BatchNorm''. After the training procedure on the digital platform, we map the pre-trained weights onto the analog computation device (simulated by the injected noise), the imprecise computing leads to severe shifts of output distribution. For implementation convenience, BatchNorm 
layer and activation function e.g., relu, are deployed on digital platform. With this advantage, we can recalculating the statistics of input mini-batch at inference phase, achieving better normalization.}
\label{fig:schematic_diagram}
\end{figure*}

\subsection{Improving Model Performance under Analog Computation}
Applying analog computation to accelerate neural network in inference phase has been an active field in recent years \cite{The_Next_Generation_of_Deep_Learning_Hardware}. Previous works improved the model performance, subject to the noise on analog devices from either the hardware level or software level. For hardware-based approaches, \cite{Statistical_Training, Efficient_precise_weight} adjusted the programming voltage to offset the conductance to compensate variation.  One disadvantage of these methods is that it is costly to read, verify and write the conductance of process-in-memory devices repeatedly.
For software (algorithm)-based approaches, 
\cite{klachko2019improving} explored the effect of common DNN (deep neural network) components and training regularization techniques, e.g., activation function, weight decay, Dropout\cite{JMLR:v15:srivastava14a}, BatchNorm, on their ability to tolerate noise on devices. \cite{qin2018training} trained a neural network with noise injected, to make the model weights less sensitive to signal variations caused by noise. \cite{zhou2020noisy} integrated the technique of knowledge distillation with noise injection training which can take advantage of the additive information of teacher model. However, as we discussed in Section~\ref{sec::intro}, these methods require prior knowledge of the noise type and need additional re-training.

\subsection{BatchNorm}

BatchNorm \cite{ioffe2015batch} is a widely used normalization technique that can accelerate and stabilize the training of a neural network by normalizing intermediate representations.
However, the effect of BatchNorm is not yet completely understood. Recent works~\cite{xie2019adversarial,frankle2020training} investigated the properties of BatchNorm to better understand its effect under different circumstances. \cite{Guo2018DoubleFP, singh2019evalnorm, summers2019things} proposed methods that mitigate the discrepancy between training and testing data, by slightly modifying the BatchNorm layers. However these methods did not consider the larger signal discrepancy caused by noise in analog computing.

%\cite{xie2019adversarial} advanced the perspective that the distribution mismatch between clean examples and adversarial examples is a key factor that causes the performance degradation in modern neural networks containing the component of BatchNorm layer, and \cite{frankle2020training} investigated the expressive power of BatchNorm by only training BatchNorm parameters which merely account for 0.5\% in the total number of model parameters. The model could only shift and rescale random features, but still achieve a fairly high accuracy. It can be seen from above previous works that the distribution shifts harm a neural net drastically, and BatchNorm is a effective way to mitigate this issue.

\section{Problem Formulation}
The progress of analog processing-in-memory is limited by the non-idealities of variations originating mainly from three factors, namely, quality of wafer manufacturing, stability of supply voltage and temperature change.  
%: 1. quality of wafer manufacturing; 2. stability of supply voltage; 3. temperature change.  
The combination of the three factors above leads to fluctuations in computation. Besides, NVM cell is a key component used in PIM-based DNN-accelerator to store the weights of neural network \cite{qin2018training} \cite{balaji2019framework}, but it suffers from the variation of electro/thermo-dynamics during the read and write operations.  In the words, the stored value in NVM cells has a tendency to fluctuate from time to time due to temperature changes and conductance drifts of the device. Moreover, these variations vary across different hardware devices, ranging from RRAM \cite{RRAM1, RRAM2, RRAM3, RRAM4}, PCRAM \cite{PCRAM} and CBRAM \cite{CBRAM}. 

According to \cite{Efficient_precise_weight}, the variations above-mentioned can be generally categorized into two types: I. MUL type and II. ADD type.  Following the approach in \cite{zhou2020noisy}, these two types of noise can be modeled using two fluctuation factors, a temporal one and a spatial one. A noise scale $\eta_0$ is used to control the severity of noise.

{\bf Temporal fluctuation}. 
In analog computing, the instability of supply voltage and the temperature rises and falls may cause different degrees of computation error. This time-varying fluctuation can be described by $N_T$ $\sim$ $\mathcal{N}$($\eta_0$,\,$\sigma_T^{2}$) which is randomly sampled for each inference batch. We also define that a noise temporal fluctuation level of 10\% means $\sigma_T$ = 0.1$\eta_0$.

{\bf Spatial fluctuation}. 
Due to the defect of the transistor manufacturing process, the analog computing variation may vary at different parts of chip, and this spatially-varying variation can be sampled from $N_S$ $\sim$ $\mathcal{N}$(1,\,$\sigma_S^{2}$) for once when the neural network is instantiated. A noise spatial fluctuation level of 10\% means that $\sigma_S$ = 0.1.

With the definitions above, the weights of noise injected model are simulated by the following two equations: MUL type variation (eq~\ref{eq:mul_type_equation}) and ADD type variation (eq~\ref{eq:add_type_equation}). 

% So the weights of model after noise injection sampled from above two types of noise sources MUL type variation $W^{mul}_{noisy}$ and ADD type variation $W^{add}_{noisy}$ will be like the following eq~\ref{eq:mul_type_equation} and eq~\ref{eq:add_type_equation}: 

\begin{equation}
W^{mul}_{noisy}= W_{orig} + W_{orig} \cdot N_T \cdot N_S
\label{eq:mul_type_equation}
\end{equation}

\begin{equation}
W^{add}_{noisy}= W_{orig} + N_T \cdot N_S
\label{eq:add_type_equation}
\end{equation}

To verify the validity of our method, we conduct all experiments with a relatively strict setting: $\sigma_T$ = 0.2$\eta_0$ and $\sigma_S$ = 0.1, at various noise scales $\eta_0$.

%temporal fluctuation level of 20\% and spatial fluctuation level of 10\% at various noise scales. 

\section{Methodology}
\label{sec::method}

\begin{algorithm2e*}
\caption{Calibrate the statistics of BatchNorm}
\label{alg:act}
Calibrated statistics of BatchNorm $\mu_{C}$ and $\sigma^2_{C}$, momentum $m$, inference data $X$, noise injected neural network $N_{noisy}$;\\
Set m $\leftarrow$ 0.999;   \tcp{A value we found suitable for datasets of different sizes.}
\For{mini-batch $x$ \textbf{in} $X$} {
\For{BN layer $B_i$ and Conv layer $L_i$ \textbf{in} $N_{noisy}$}{
Calculate mean $\mu$ and variance $\sigma^2$ of mini-batch activations $L_i$($x$) and update the statistics of $B_i$;\\

\If{$\mu_{Ci}$ and $\sigma_{Ci}$ is not initialized}{
Initialize $\mu_{Ci}$ and $\sigma^2_{Ci}$ as the statistics of first mini-batch activations;\\
}
\Else{
$\mu_{Ci}$ $\leftarrow$ $m$ $\cdot$ $\mu_{Ci}$ + (1 - $m$) $\cdot$ $\mu$;\\
$\sigma^2_{Ci}$ $\leftarrow$ $m$ $\cdot$ $\sigma^2_{Ci}$ + (1 - $m$) $\cdot$ $\sigma^2$;
}
}
}
\end{algorithm2e*}

\subsection{Inference with Noise-Aware BatchNorm}
Traditional BatchNorm contains two statistical and two learnable components: mean, variance, scale and bias. In the training phase, BatchNorm calculates the mean $E[x]$ and variance $Var[x]$ of mini-batch $x$, and then normalizes each scalar feature independently to zero mean and unit variance. After that, BatchNorm scales and shifts the normalized values $\hat{x}$ by the learnable parameters, scale $\gamma$ and bias $\beta$. Meanwhile, BatchNorm also maintains the exponential moving average (EMA) of the mini-batch mean and the mini-batch variance which can represent the training data distribution to normalize the input batch during testing phase.

\begin{equation}
\hat{x}= \frac{x - E[x]}{\sqrt{Var[x] + \epsilon}} \cdot \gamma + \beta
\label{eq:batchnorm}
\end{equation}

To model the analog computation noise, the noise is injected into model weights, making the internal activations distribution greatly differ from the original activations of the clean weights. However, traditional BatchNorm performs the normalization with EMA of training data without considering noise injection. Such EMA can not successfully normalize the noisy activations, and therefore reduces the effectiveness of BatchNorm in adjusting the activations distribution of previous layer. As a result, the computation error caused by noisy weights keeps propagating, and eventually leads to wrong predictions.

To mitigate the distribution shift between the noisy activations and original ones, we conjecture a perspective that differs from the traditional paradigm that we should not use the statistical results maintained in training phase when there is an obvious distribution shift. We propose a simple yet effective method called \emph{``Noise-Aware BatchNorm''} and illustrate its main idea in Figure~\ref{fig:schematic_diagram}. By recalculating the mini-batch statistics with the noise injected into the model, the statistic of such unstable analog computing environment can be tracked without knowing the actual characteristics of injected noise. The calibrated statistics normalize the intermediate activations properly, and effectively pull the distribution of noisy activations closer to that of the clean ones. 

Specifically, to adapt a previously trained model to noise, our proposed method first performs BatchNorm calibration over the training data with noise injected to the model weights. Algorithm~\ref{alg:act} outlines the steps of this calibration process to compute the calibrated mean and variance for each BatchNorm layer. The final calibrated mean and variance values will be used to perform normalization at testing time.
We want to note that our proposed calibration is done on training data, that is, no prior knowledge of the testing data is incorporated when computing the calibrated mean and variance.  

At testing time, while one choice is to fix the calibrated means and variances as constants when processing the test data, our method also allows us to keep tracking the characteristics of noise by continuing the calibration process whenever the model has processed a batch of test data. This property of \textit{dynamic calibration} at the testing time shows good adaptability to changes in the noise environment, which is demonstrated in Section ~\ref{exp:Comparison_with_Noise_Injection_Training}.

\section{Experiments}

\subsection{Image Classification}

\begin{table*}[htbp]
\tabcolsep=10pt
\caption{Validation acc (\%) on ImageNet-2012 Under MUL Type Variation.}
\begin{center}
\begin{tabular}{|c|c|c|c|c|c|}
\hline
\multirow{2}{*}{Model}
&\multicolumn{5}{|c|}{Noise Scale} \\
\cline{2-6} 
&$\eta_0$ = 0.02 &$\eta_0$ = 0.04 &$\eta_0$ = 0.06 &$\eta_0$ = 0.08 &$\eta_0$ = 0.10 \\
\hline
%  &73.31
% 72.296
ResNet-34 \cite{he2015deep} &71.53 &65.73 &49.92 &20.55 &3.66\\
ResNet-34+NaBN &72.34 &72.28 &72.07 &71.97 &71.64 \\
\hline
% &76.13
% 74.810
ResNet-50 \cite{he2015deep}  &72.85 &56.88 &19.49 &2.73 &0.40\\
ResNet-50+NaBN &74.81 &74.79 &74.71 &74.57 &74.54\\
\hline
%  &77.37
% 76.310
ResNet-101 \cite{he2015deep} &73.80 &52.89 &9.32 &0.38 &0.08\\
ResNet-101+NaBN &76.30 &76.27 &76.25 &76.12 &75.94\\
\hline
% &78.47
% 76.098
WideResNet-50-2 \cite{zagoruyko2016wide}  &75.09 &59.24 &23.36 &3.93 & 0.59\\
WideResNet-50-2+NaBN &76.10 &76.08 &75.95 &75.77 &75.72\\
\hline
% &71.88
% 69.034
MobileNet-v2 \cite{s2018mobilenetv2} &49.61 &1.96 &0.19 &0.13 &0.12\\
MobileNet-v2+NaBN &68.76 &67.97 &66.90 &65.26 &63.41\\
\hline
% \multicolumn{4}{l}{$^{\mathrm{a}}$Sample of a Table footnote.}
\end{tabular}
\label{classification_mul_type}
\end{center}
\end{table*}

\begin{table*}[htbp]
\tabcolsep=10pt
\caption{Validation acc (\%) on ImageNet-2012 Under ADD Type Variation.}
\begin{center}
\begin{tabular}{|c|c|c|c|c|}
\hline
\multirow{2}{*}{Model}
&\multicolumn{4}{|c|}{Noise Scale} \\
\cline{2-5} 
&$\eta_0$ = 0.0002 &$\eta_0$ = 0.00025 &$\eta_0$ = 0.0003 &$\eta_0$ = 0.00035\\
\hline
% 72.296
ResNet-34 \cite{he2015deep} &61.77 &46.28 &26.88 &10.83\\
ResNet-34+NaBN &72.36 &72.31 &72.18 &72.08\\
\hline
% 74.810
ResNet-50 \cite{he2015deep} &61.64 &45.85 &24.64 &9.46\\
ResNet-50+NaBN &74.93 &74.92 &74.90 &74.86\\
\hline
% 76.310
ResNet-101 \cite{he2015deep} &39.35 &9.83 &1.21 &0.25\\
ResNet-101+NaBN &76.44 &76.39 &76.28 &76.10\\
\hline
% 76.098
WideResNet-50-2 \cite{zagoruyko2016wide} &42.45 &14.69 &3.31 &0.86\\
WideResNet-50-2+NaBN &76.07 &75.93 &75.79 &75.62\\
\hline
% 69.034
MobileNet-v2 \cite{s2018mobilenetv2} &67.23 &64.27 &60.21 &54.97\\
MobileNet-v2+NaBN &68.85 &68.77 &68.75 &68.66\\
\hline
% \multicolumn{4}{l}{$^{\mathrm{a}}$Sample of a Table footnote.}
\end{tabular}
\label{classification_add_type}
\end{center}
\end{table*}

% In this section, experiments are conducted with widely-used dataset ImageNet-2012 \cite{krizhevsky2012imagenet} with different architectures. And we show the performance degradation when the noise is injected into model weights, and with the proposed method, the problem of distribution mismatch is eased, leads to favorable performance.

\textbf{Dataset and model.} Our experiments on image classification are conducted with the official PyTorch pre-trained ResNet \cite{he2015deep, zagoruyko2016wide} and MobileNet-v2 \cite{s2018mobilenetv2} models, on the ImageNet-2012 \cite{krizhevsky2012imagenet} dataset. ImageNet-2012 is a large-scale dataset containing roughly 1.3 million training images and 50k validation images of 1k categories.

\textbf{Results.}
Our method demonstrates improvements over all of the state-of-the-art  models compared to baseline, improving validation accuracy by a significant margin (Tables~\ref{classification_mul_type} and~\ref{classification_add_type}).
While the performance of vanilla models degrades when the noise is injected into model weights, our proposed method \emph{``Noise-Aware BatchNorm''} eases the problem of distribution shifts, leading to favorable performance.

Besides, we find that the deeper the network is, the more susceptible it is to noise. There are obvious performance gaps of ResNet family at different depths, the accuracy of ResNet-34, 50, 101 are 49.92\%, 19.49\%, 9.32\% respectively under the noise scale of 0.06. We believe that it is due to the propagation of noise.

When it comes to the width of architecture, previous work suggests that increasing the width of the network is regarded to be more robust to variations such as adversarial examples \cite{gao2019convergence}, due to the larger amount of parameters in the wider models. However, comparing the results of ResNet-50 and WideResNet-50-2 in our experiments, the width of a model does not seem to have any benefit in terms of noise resistance in analog computing circumstance. 
We conjecture that any additional parameters in the wider model will be affected by the noise as well, when deployed on processing-in-memory device. 
% And there is another interesting found that MobileNet-v2 seems to be more robust to ADD type variation, but have poor resistance to MUL type variation.

\label{sec::experiments}
\begin{table*}[htbp]
\tabcolsep=10pt
\caption{Validation mAP, mAR (\%) at 0.5 IoU on COCO-2014 val 5k Under MUL/ADD Type Variation.}
\begin{center}
\begin{tabular}{|c|c|c|c|c|c|c|c|c|}
\hline
\multirow{3}{*}{Model}
&\multicolumn{4}{|c|}{MUL Type Vatiation}
&\multicolumn{4}{|c|}{ADD Type Vatiation}
% &\multicolumn{4}{|c|}{Noise Scale}
\\
\cline{2-9} 
&\multicolumn{2}{|c}{$\eta_0$ = 0.05}
&\multicolumn{2}{|c|}{$\eta_0$ = 0.10}
&\multicolumn{2}{|c|}{$\eta_0$ = 0.00035}
&\multicolumn{2}{|c|}{$\eta_0$ = 0.00045}
\\
\cline{2-9} 
&\multicolumn{1}{|c}{mAP} &\multicolumn{1}{|c}{mAR}
&\multicolumn{1}{|c}{mAP} &\multicolumn{1}{|c}{mAR}
&\multicolumn{1}{|c}{mAP} &\multicolumn{1}{|c}{mAR}
&\multicolumn{1}{|c}{mAP} &\multicolumn{1}{|c|}{mAR}
\\

\hline
YOLO-v3 \cite{redmon2018yolov3} &30.5 &58.0 &0.1 &1.6 &39.3 &65.0 &26.4 &54.9 \\
\hline
YOLO-v3+NaBN &49.8 &73.5 &48.7 &71.8 &47.8 &73.1 &43.1 &71.0\\
\hline

\end{tabular}
\label{object_detection_mul_type}
\end{center}
\end{table*}

\begin{table*}[htbp]
\tabcolsep=10pt
\caption{Validation mIoU, PixAcc (\%) on PASCAL VOC-2012 Under MUL/ADD Type Variation.}
\begin{center}
\begin{tabular}{|c|c|c|c|c|c|c|c|c|}
\hline
\multirow{3}{*}{Model}
&\multicolumn{4}{|c|}{MUL Type Variation}
&\multicolumn{4}{|c|}{ADD Type Variation}
\\
\cline{2-9} 
% &\multicolumn{8}{|c|}{Noise Scale}
% \\
% \cline{2-9} 
&\multicolumn{2}{|c}{$\eta_0$ = 0.05}
&\multicolumn{2}{|c|}{$\eta_0$ = 0.10}
&\multicolumn{2}{|c|}{$\eta_0$ = 0.00015}
&\multicolumn{2}{|c|}{$\eta_0$ = 0.00025}
\\
\cline{2-9} 
&\multicolumn{1}{|c}{mIoU} &\multicolumn{1}{|c}{PixAcc}
&\multicolumn{1}{|c}{mIoU} &\multicolumn{1}{|c}{PixAcc}
&\multicolumn{1}{|c}{mIoU} &\multicolumn{1}{|c}{PixAcc}
&\multicolumn{1}{|c}{mIoU} &\multicolumn{1}{|c|}{PixAcc}
\\

\hline
FCN-ResNet101 \cite{long2014fully} &70.8 &83.8 &43.1 &62.5 &65.5 &80.3 &58.2 &75.0\\
\hline
FCN-ResNet101+NaBN  &80.6 &89.9 &79.7 &89.3 &80.9 &90.0 &74.4  &86.1\\
\hline

% \multicolumn{4}{l}{$^{\mathrm{a}}$Sample of a Table footnote.}
\end{tabular}
\label{segmentation_mul_type}
\end{center}
\end{table*}

\subsection{Object Detection \& Semantic Segmentation}
\textbf{Dataset and model.}
For the task of object detection, we use the official pretrained YOLO-v3 \cite{redmon2018yolov3} implementation as our detector, and conduct experiments on COCO-2014 dataset \cite{lin2014microsoft} which has 80 object classes. %Our detector is trained on trainval35k set with around 75k images, and evaluate on a held out set with 5k images. 
We use the standard COCO evaluation metrics of mean average precision (mAP) and mean average recall (mAR), with IoU threshold set at 0.5.

As for the task of semantic segmentation, we use the PyTorch official FCN-ResNet101 \cite{long2014fully} that is pretrained on part of COCO dataset which shares the same classes with PASCAL VOC-2012 \cite{pascal-voc-2012}. We experiment on PASCAL VOC-2012 with 21 classes including background, and use the pixel accuracy (PixAcc) and mean Intersection of Union (mIoU) as the evaluation metrics with single scale evaluation. Specifically, we follow the procedure in the standard competition benchmark and calculate mIoU by ignoring the pixels that are labeled as ``background''.

\textbf{Results.}
Tables~\ref{object_detection_mul_type} and~\ref{segmentation_mul_type} show the results on the object detection and the semantic segmentation, respectively.  The results demonstrate the effectiveness of our method on different computer vision tasks compared to vanilla models.

%The results are shown in Table~\ref{object_detection_mul_type}, Table~\ref{segmentation_mul_type} repectively, which demonstrate the effectiveness of our method on different computer vision tasks compared to vanilla models.

\subsection{Comparison with Noise Injection Training}
\label{exp:Comparison_with_Noise_Injection_Training}
Noise Injection Training \cite{qin2018training} was first proposed to improve robustness against noisy computation during inference for RNNs models, but the concept can be applied to CNNs models too. Its main idea is that, by learning the distribution of injection noise, a model can avoid performance degradation on various hardware devices at inference time. Though Noise Injection Training achieves favorable performance under the specific noise it trained on, it requires prior knowledge of the noise beforehand, which is unrealistic in practice.  

To show the adaptability of our method, we compare our method with Noise Injection Training, under a circumstance that the noise characteristic varies.

\textbf{Dataset and model.}
We use PyTorch official pre-trained ResNet-50 to conduct experiments on ImageNet-2012 under the injection of MUL type noise. %In order to investigate the property when the noise of computation platform varies, 
The noise injection trained model initializes its weights as pre-trained model, fine-tunes at noise scale $\eta_0$ of 0.06 with an initial learning rate of 0.001 and cosine learning rate scheduling until the validation accuracy achieves the pre-trained baseline accuracy. For our proposed method, we calibrate pre-trained model at noise scale $\eta_0$ of 0.06 on training set, with ``dynamic calibration'' at the testing time (Section~\ref{sec::method}).  
%and the transfer noise scale on testing set.

\begin{figure}[htbp]
\centering
\includegraphics[width=0.46\textwidth]{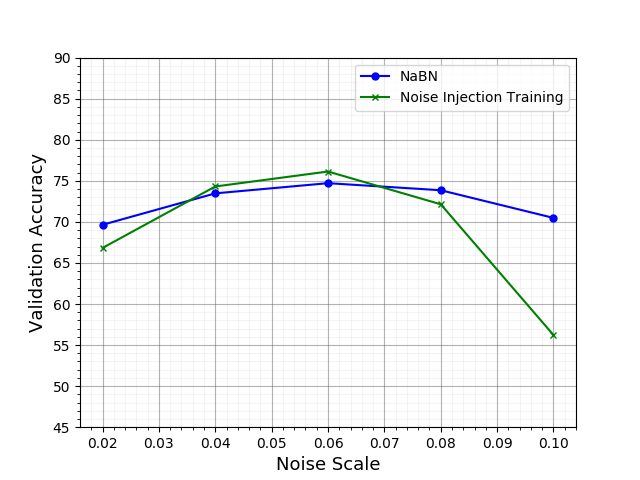}
\caption{Comparison of the validation accuracy of ImageNet-2012 under MUL type variation between Noise-Aware BatchNorm and Noise Injection Training.}
\label{fig:compare_noise_injection_training}
\end{figure}

\textbf{Results.}
As shown in Figure ~\ref{fig:compare_noise_injection_training}, when the noise scale varies away from the injected noise scale ($\eta_0$=0.06), the performance of the noise injection model drops dramatically, showing the model overfits on specific noise. However, our dynamic calibrated model has better adaptability to various scales of noise without further re-training and prior information of noise, which is a favorable property for complex analog computation environments where the characteristic of noise may vary over time.

%Dynamic calibration.

\section{Conclusion}
\label{sec::conclusion}
We investigate a concept that formulates the imprecise processing-in-memory computing as a distribution shift problem, and propose an effective yet simple way to adjust the mismatching distribution by a novel calibration of statistics in BatchNorm layers. We conduct extensive experiments on important vision tasks including classification, object detection and semantic segmentation. 
Experimental results demonstrate the effectiveness and generalizability of our method, showing that it is a promising solution to facilitate the progress of neural network in analog computing.

%We also compare the effectiveness and generalizability to previous method, the results demonstrate the significant improvement, further facilitate the progress the development of neural network in the analog computing field.

% \clearpage

\vspace{12pt}

\end{document}